\documentclass[letterpaper, 10 pt, conference]{ieeeconf}  %

\IEEEoverridecommandlockouts                              %
\usepackage{amsmath,amssymb,amsfonts}
\usepackage{bm,bbm,siunitx,hyperref,graphicx}
\usepackage{float}	%
\usepackage{xcolor}
\usepackage{mathtools, nccmath}

\let\labelindent\relax%

\usepackage{enumitem}
\usepackage{makecell}
\usepackage{cellspace}
\usepackage{cite}
\usepackage{balance}   %

\DeclareSIUnit \ampereHour {Ah}

\newcommand{\smallHalf}{\ensuremath{\tfrac{1}{2}}}

\newcommand{\mmat}[1]{\begin{bmatrix} #1 \end{bmatrix} }

\newcommand{\mrb}[1]{\left( #1 \right)} %

\newcommand{\commentOut}[1]{}

\newcommand{\figref}[1]{Fig.~\ref{#1}}
\newcommand{\secref}[1]{Section~\ref{#1}}

\newcommand{\bs}[1]{\boldsymbol{#1}}
\newcommand{\sss}{\scriptscriptstyle}

\newcommand{\earthFrame}{\mathrm{E}}
\newcommand{\mainQuadFrame}{\mathrm{Q}}
\newcommand{\cameraFrame}{\mathrm{C}}
\newcommand{\flyingBatteryFrame}{\mathrm{F}}
\newcommand{\markerFrame}{\mathrm{M}}

\newcommand{\E}{\earthFrame}
\newcommand{\Q}{\mainQuadFrame}
\newcommand{\C}{\cameraFrame}
\newcommand{\F}{\flyingBatteryFrame}
\newcommand{\M}{\markerFrame}

\newcommand{\rotation}[2]{\mathbf{R}^{\sss #1 #2}}
\newcommand{\translation}[3]{\mathbf{s}_{\sss #1 #2}^{\sss #3}}
\newcommand{\R}{\rotation}
\newcommand{\T}{\translation}
\newcommand{\position}{\translation}

\newcommand{\velocity}[3]{\dot{\mathbf{s}}_{\sss #1 #2}^{\sss #3}}
\newcommand{\acceleration}[3]{\ddot{\mathbf{s}}_{\sss #1 #2}^{\sss #3}}

\newcommand{\rotationRate}[2]{\dot{\mathbf{R}}^{\sss #1 #2}}
\newcommand{\angVel}[1]{\bs{\omega}^{\sss #1}}
\newcommand{\angAcc}[1]{\dot{\bs{\omega}}^{\sss #1}}

\newcommand{\gravity}[1]{\mathbf{g}^{\sss #1}}

\newcommand{\skewMat}[1]{\mathbf{S} \! \mrb{#1} \!}
\newcommand{\expS}[1]{\exp \! \mrb{\skewMat{#1}} \!}
\newcommand{\rotInv}[1]{\mathrm{rot}^{-1} \! \mrb{#1} \!}
\newcommand{\TMat}[1]{\mathbf{T} \! \mrb{#1} \!}

\newcommand{\camAttError}{\widetilde{\bs{\sigma}}}
\newcommand{\transpose}[1]{{#1}^T}

\newcommand{\realVecs}[1]{\ensuremath{\mathbb{R}^{#1}}}
\newcommand{\SOThree}{\mathrm{SO}(3)}
\newcommand{\soThree}{\mathfrak{so}(3)}
\newcommand{\realMats}[2]{\ensuremath{\mathbb{R}^{#1 \times #2}}}

\newcommand{\mass}[1]{m_{\mathrm{\sss #1}}}
\newcommand{\inertiaMat}[1]{\mathbf{J}_{\mathrm{\sss #1}}}
\newcommand{\forces}[1]{\mathbf{f}^{\mathrm{\sss #1}}}
\newcommand{\torques}[1]{\bs{\tau}^{\mathrm{\sss #1}}}

\newcommand{\accelerometer}[1]{\widetilde{\mathbf{a}}}
\newcommand{\rateGyro}[1]{\widetilde{\bs{\omega}}}
\newcommand{\noiseAccel}[1]{\bs{\nu}_{\mathbf{a}}}
\newcommand{\noiseGyro}[1]{\bs{\nu}_{\bs{\omega}}}
\newcommand{\noiseVarAccel}{\sigma_\mathbf{a}}
\newcommand{\noiseVarGyro}{\sigma_{\bs{\omega}}}

\newcommand{\cameraPos}{\widetilde{\T{\M}{\C}{\C}}}
\newcommand{\cameraAtt}{\widetilde{\R{\M}{\C}}}
\newcommand{\noiseCameraPos}{\bs{\nu}_{\mathbf{s}}}
\newcommand{\noiseCameraAtt}{\bs{\nu}_{\mathbf{\sss R}}}
\newcommand{\noiseCovCameraPos}{\bs{\Sigma}_\mathbf{s}}
\newcommand{\noiseCovCameraAtt}{\bs{\Sigma}_{\sss \mathbf{R}}}

\newcommand{\cameraAttRef}{\widehat{\R{\M}{\C}_{\mathrm{ref}}}}

\newcommand{\relState}{\widehat{\mathbf{x}}}
\newcommand{\relStatePos}{\widehat{\position{\Q}{\F}{\E}}}
\newcommand{\relStateVel}{\widehat{\velocity{\Q}{\F}{\E}}}
\newcommand{\absStateAttErr}{\widehat{\bs{\delta}^{\sss \E \Q}}}
\newcommand{\absStateAttRef}{\widehat{\R{\E}{\Q}_{\mathrm{ref}}}}
\newcommand{\absStateAtt}{\widehat{\R{\E}{\Q}}}

\newcommand{\relStatePosEasy}{\widehat{\mathbf{p}}}
\newcommand{\relStateVelEasy}{\widehat{\mathbf{v}}}

\newcommand{\absStateAttErrEasy}{\widehat{\bs{\delta}}}

\newcommand{\tNow}{(t)}
\newcommand{\tPlus}{(t+\Delta t)}
\newcommand{\deltaT}{\Delta t}

\newcommand{\covRelState}{\mathbf{P}_{\mathbf{xx}}}
\newcommand{\covRelStatePred}{\mathbf{P}_{\mathbf{xx},p}}
\newcommand{\covRelStatePredPlus}{\mathbf{P}_{\mathbf{xx},p_+}}
\newcommand{\covRelStateMeas}{\mathbf{P}_{\mathbf{xx},m}}

\newcommand{\identity}{\mathbf{I}}
\newcommand{\zeroMat}{\mathbf{0}}

\newcommand{\HMat}{\mathbf{H}\tNow}
\newcommand{\KMat}{\mathbf{K}\tNow}

\newcommand{\innovation}{\mathbf{e}_m}

\title{\LARGE \bf
Docking two multirotors in midair using relative vision measurements
}

\author{
Karan P. Jain, Minos Park, and Mark W. Mueller%
\thanks{The authors are with the High Performance Robotics Lab (HiPeRLab) at the Dept. of Mechanical Engineering, UC Berkeley, CA 94720, USA.
{\tt\small \{karanjain, minos.park, mwm\}@berkeley.edu}} }%

\begin{document}

\maketitle

\begin{abstract}
	Modular robots have been rising in popularity for a variety of applications, and autonomous midair docking is a necessary task for real world deployment of these robots.
	We present a state estimator based on the extended Kalman filter for relative localization of one multirotor with respect to another using only onboard sensors, specifically an inertial measurement unit and a camera-marker pair.
	Acceleration and angular velocity measurements along with relative pose measurements from a camera on the first multirotor looking at a marker on the second multirotor are used to estimate the relative position and velocity of the first multirotor with respect to the second, and the absolute attitude of the first multirotor.
	We also present a control architecture to use these onboard state estimates to control the first multirotor at a desired setpoint with respect to the second.
	The performance of the estimator and control architecture are experimentally validated\footnote{The explanation and experimental validation video can be found here: \url{https://youtu.be/m9YqOm3VtTM}} by successfully and repeatably performing midair docking -- a task that requires relative position precision on the order of a centimeter.
\end{abstract}

\section{Introduction}\label{sec:intro}
Modular and cooperative flying robots have started seeing increasing use in applications such as lifting objects \cite{duffy2015lift, gabrich2018flying}, assisting each other such as for flight time extension \cite{jain2020flying}, or demonstrating aesthetic aerial transformation \cite{zhao2018design}.
These multiple small robots offer several advantages over a few larger robots.
The individual units are usually simple in design.
The failure of a few of these units does not affect the high-level objective, offering redundancy.
Lastly, they can move through tight, cluttered spaces easily as individual units and then group up at the desired location.

A critical aspect of having these modular robots to work in real life settings is autonomous assembly and dismantling.
For flying robots, this translates to autonomous midair docking and undocking.
Current literature explores docking via various methods such as using a robotic arm and winch \cite{miyazaki2018airborne}, using magnets \cite{li2019modquad}, or using mechanical structures, \cite{jain2020flying, rocha2020toward}.

Midair docking is a sensitive task that requires a coarse global localization to roughly get the vehicles close to each other and a fine relative localization for the actual docking task.
In this paper, we focus on the precise relative localization aspect.
A combination of inertial measurement units and computer vision based sensors is usually the most popular choice for this task.
For example, it has been used for autonomous relative navigation or docking of aircrafts \cite{johnson2007real, wilson2015guidance}, relative position sensing for an underwater vehicle \cite{huster2003relative}, and even considered for docking spacecrafts \cite{kelsey2006vision, regoli2014line}.

\begin{figure}
	\centering
	\includegraphics[width=0.92\columnwidth]{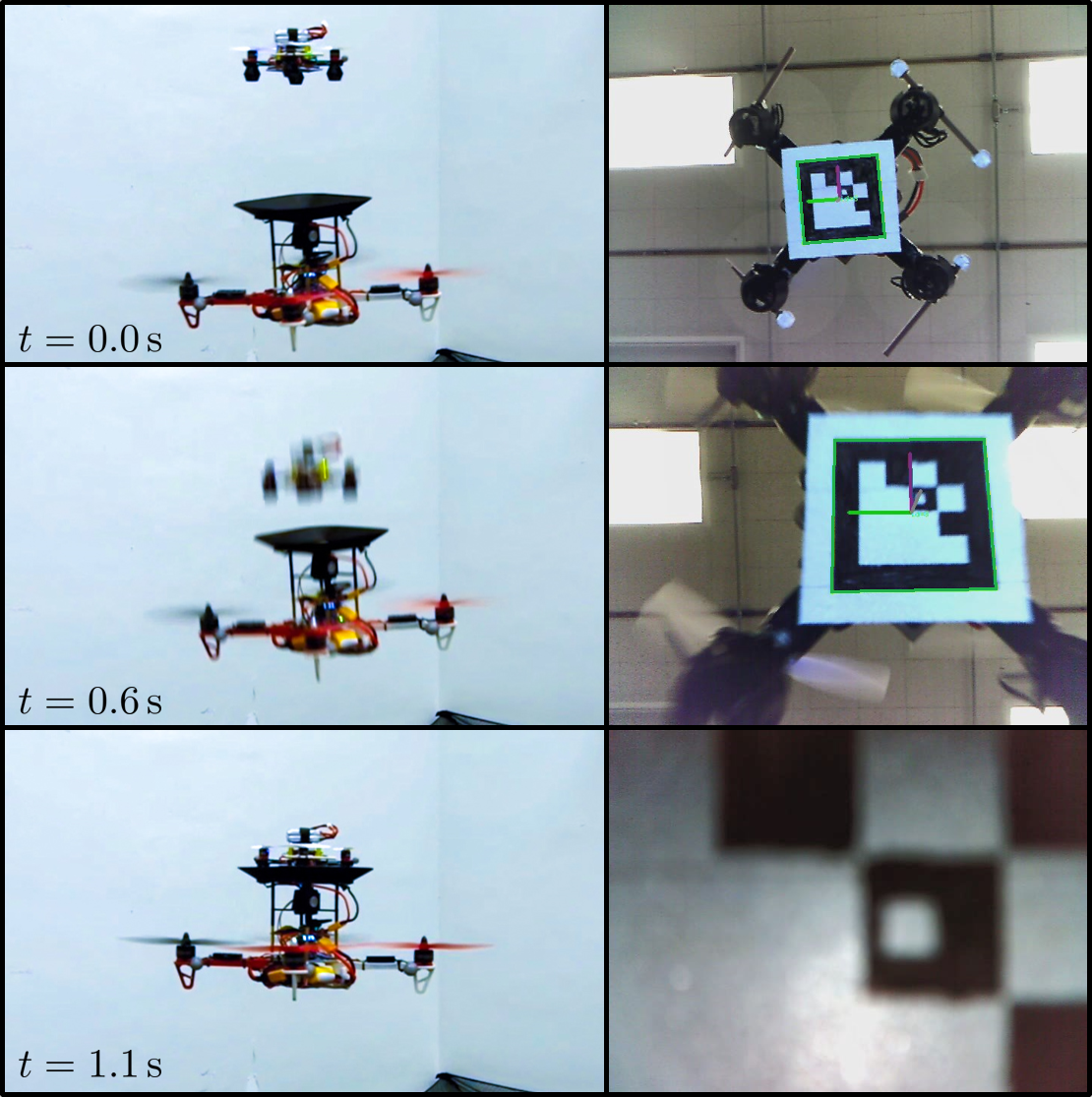}
	\vspace{-2.5mm}
	\caption{
		Offboard (left) and onboard (right) sequence of images of the docking procedure.
		From top to bottom:
		(i) the two quadcopters flying in proximity;
		(ii) passive quadcopter is commanded to stop its motors to dock since it is within docking range;
		(iii) the two quadcopters are docked.
	}
	\vspace{-8mm}
	\label{fig:FilmReelDocking}
\end{figure}

In this work, we present a complete 6 degree of freedom (DoF) localization of one multirotor with respect to another, using only onboard sensors.
Specifically, we derive a state estimation algorithm based on the extended Kalman filter (EKF) that takes, as input, inertial measurements (acceleration and angular velocity) and relative pose (position and orientation) measurements.
These measurements are then used to estimate the relative position and velocity of one multirotor with respect to another, and its absolute attitude with respect to the inertial frame.
We also present a control architecture to experimentally validate the relative localization algorithm using an onboard camera on the first quadcopter and a marker (which provides full 6 DoF information on detection) on the second, and control the first quadcopter using just the onboard estimates.
Lastly, we successfully demonstrate midair docking using these onboard estimates -- a task that requires a position precision on the order of a centimeter.
\figref{fig:FilmReelDocking} shows pictures of two quadcopters during a docking maneuver where the lower quadcopter is using purely onboard sensing.

The paper is organized as follows.
\secref{sec:systemDynamics} covers basic multirotor dynamics and the architecture of our control system.
In \secref{sec:relStateEst}, we derive the equations for the relative state estimation algorithm using the EKF formulation.
\secref{sec:hardwareDesign} explains the design and selection of various hardware components used in our experiments.
Lastly, in \secref{sec:expValidation}, we experimentally validate the state estimator and present results from successful midair docking experiments.
\section{System Dynamics and Architecture}\label{sec:systemDynamics}
In this section, we briefly describe the dynamics of a multirotor, and provide details about the architecture of the system.
We introduce the relevant states of the system and their governing equations, and how we can use onboard sensors to estimate these states.
The dynamics that we show here are specific to a multirotor, but the estimation algorithm that we present is indifferent to these dynamics and can be applied to any aerial vehicle.

\subsection{Notation}\label{sec:notation}
We represent scalars with non-bold symbols such as $m$, vectors with lowercase bold symbols such as $\gravity{}$, and matrices with uppercase bold symbols such as $\R{}{}$.
A vector $\mathbf{v}$ in a certain frame $\mathrm{F}$ is denoted by $\mathbf{v}^\mathrm{\sss F}$.

One can go from any frame to any other frame via a rigid body transformation -- a rotation and a translation.
Rotations from frame $\mathrm{F_1}$ to frame $\mathrm{F_2}$ are represented by the rotation matrix $\R{\mathrm{F_2}}{\mathrm{F_1}}$.
To transform a vector from one frame to another, the following relation is used
\begin{equation}
	\mathbf{v}^{\sss \mathrm{F_2}} = \R{\mathrm{F_2}}{\mathrm{F_1}} \mathbf{v}^{\sss \mathrm{F_1}} .
\end{equation}

The displacement vector going from origin of frame $\mathrm{F_1}$ to origin of frame $\mathrm{F_2}$ is given by $\T{\mathrm{F_2}}{\mathrm{F_1}}{}$. This vector can be represented in any frame of our choice, say $\mathrm{F}$, as $\T{\mathrm{F_2}}{\mathrm{F_1}}{\mathrm{F}}$.

\subsection{Quadcopter Dynamics}
A typical multirotor can be modeled as a rigid body with six degrees of freedom.
The body-fixed frame is denoted by $\Q$.
The inertial frame of reference (or Earth frame) is denoted by $\E$.
The position, velocity, and acceleration of this multirotor with respect to the inertial frame are denoted by $\position{\Q}{\E}{}, \velocity{\Q}{\E}{}, \acceleration{\Q}{\E}{} \in \realVecs{3}$.

The attitude of the multirotor is represented by the rotation matrix $\R{\E}{\Q} \in \SOThree$.
It evolves as a function of the angular velocity $\angVel{\Q}$ and angular acceleration $\angAcc{\Q}$,
\begin{equation}
	\rotationRate{\E}{\Q} = \R{\E}{\Q} \skewMat{\angVel{\Q}} \\
\end{equation}
where $\skewMat{\cdot} \colon \realVecs{3} \to \soThree$ produces the skew-symmetric form of the vector argument, and $\soThree$ represents the Lie algebra of the rotation group.
Specifically, if $\mathbf{a} = (a_1, a_2, a_3)$ then
\begin{equation}
	\skewMat{\mathbf{a}} = \mmat{
		   0	& -a_3	&  a_2 \\ 
		 a_3	&    0	& -a_1  \\
		-a_2	&  a_1	&    0
	}.
\end{equation}

The external forces and torques acting on the system directly relate to the translational and rotational accelerations, and can be expressed via the Newton-Euler equations,
\begin{equation}
\begin{aligned}
	\acceleration{\Q}{\E}{} &= \frac{\forces{}}{\mass{\Q}} + \gravity{} \\
	\angAcc{\Q}  &= \inertiaMat{\Q}^{-1} \mrb{\torques{\Q} - \skewMat{\angVel{\Q}} \inertiaMat{\Q} \angVel{\Q}}
\end{aligned}
\label{eq:dynamics}
\end{equation}
where $\mass{\Q}, \inertiaMat{\Q}$ are the multirotor's mass and (body-fixed) moment of inertia respectively, $\forces{}$ is the sum of forces acting on the multirotor (except weight due to gravity $\gravity{}$), and $\torques{Q}$ is the sum of external torques acting on the vehicle (expressed in the multirotor's frame).

The total force $\forces{}$ can be broken down into the propeller forces $\forces{}_p$ and disturbance forces $\forces{}_d$.
Similarly, the total torques $\torques{}$ are a result of moments produced by the propellers $\torques{}_p$ and disturbance torques $\torques{}_d$.
The disturbance forces and torques are usually hard to characterize and heavily depend upon the application in which the multirotor is being used.
They might also come up due to modeling errors, aerodynamic effects, misaligned actuators, among other factors.
In our case, we have two multirotors flying in proximity which results in aerodynamic interference as analyzed in \cite{jain2019modeling}.

\subsection{System Architecture}
\begin{figure}
	\centering
	\includegraphics[width=0.9\columnwidth]{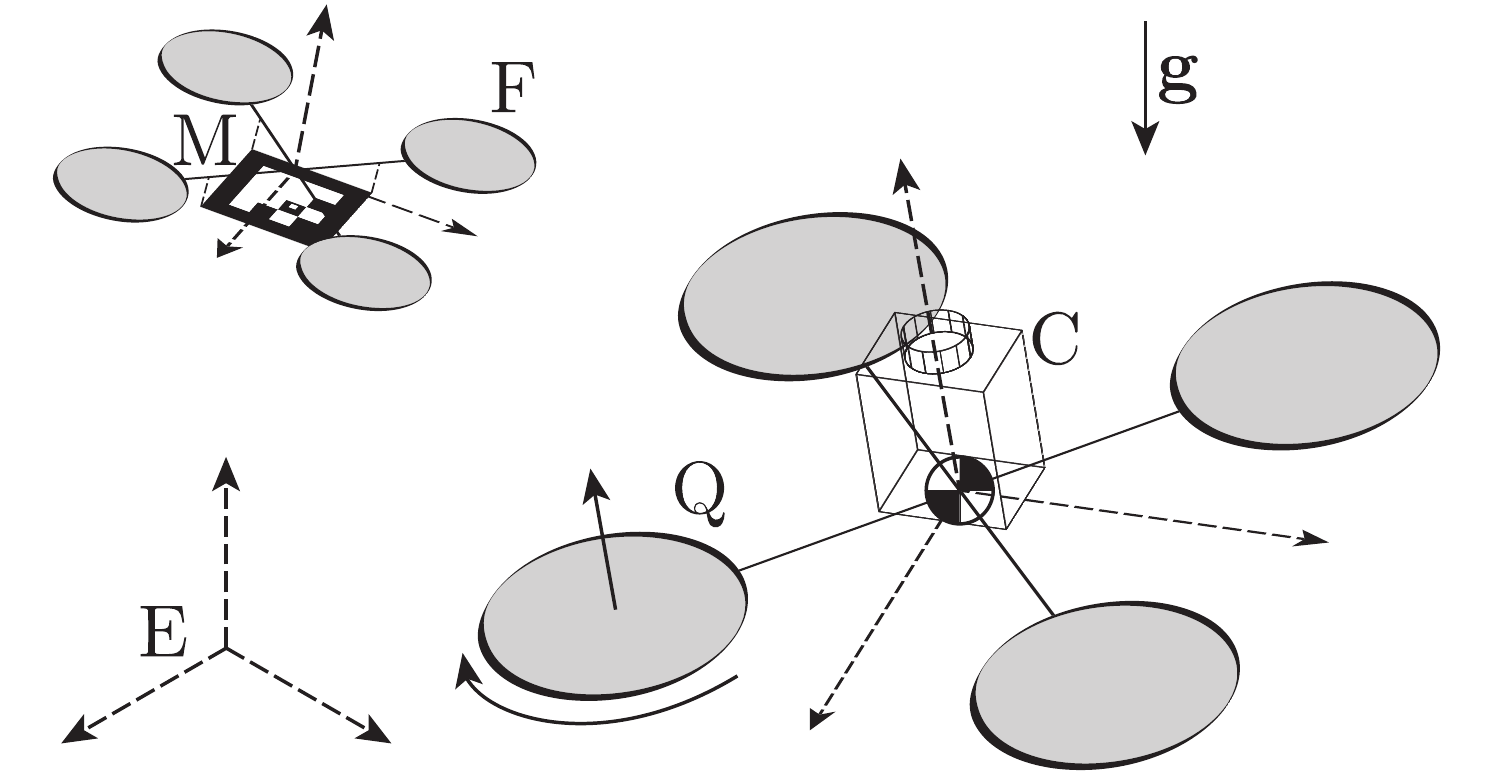}
	\vspace{-1mm}
	\caption{Schematic of a docking flight and relevant reference frames}
	\vspace{-7mm}
	\label{fig:TwoQuadCartoon}
\end{figure}

We describe the estimation and control architecture for docking two quadcopters midair using relative pose measurements from a camera looking at a marker.
We have a ``passive'' quadcopter which is supposed to hover in one place, carrying a marker with it that can be seen by the ``active'' quadcopter.
The active quadcopter carries a camera and its objective is to localize itself with respect to the passive quadcopter so that it can dock with it.
\figref{fig:TwoQuadCartoon} shows a diagram of a typical flight during midair docking, along with all the frames of reference.

There are five frames of interest -- 
the inertial or Earth-fixed frame $\E$,
the frame fixed to the active quadcopter $\Q$,
the camera frame (mounted on the active quadcopter) $\C$,
the frame fixed to the passive quadcopter $\F$,
and the marker frame (pasted on the passive quadcopter) $\M$.

Whenever the passive quadcopter's marker is not in the Field of View (FoV) of the active quadcopter's camera, both the quadcopters are localized using sensor fusion of a motion capture system and an onboard rate gyroscope \cite{jain2020staging}.

\begin{figure}
	\centering
	\includegraphics[width=0.9\columnwidth]{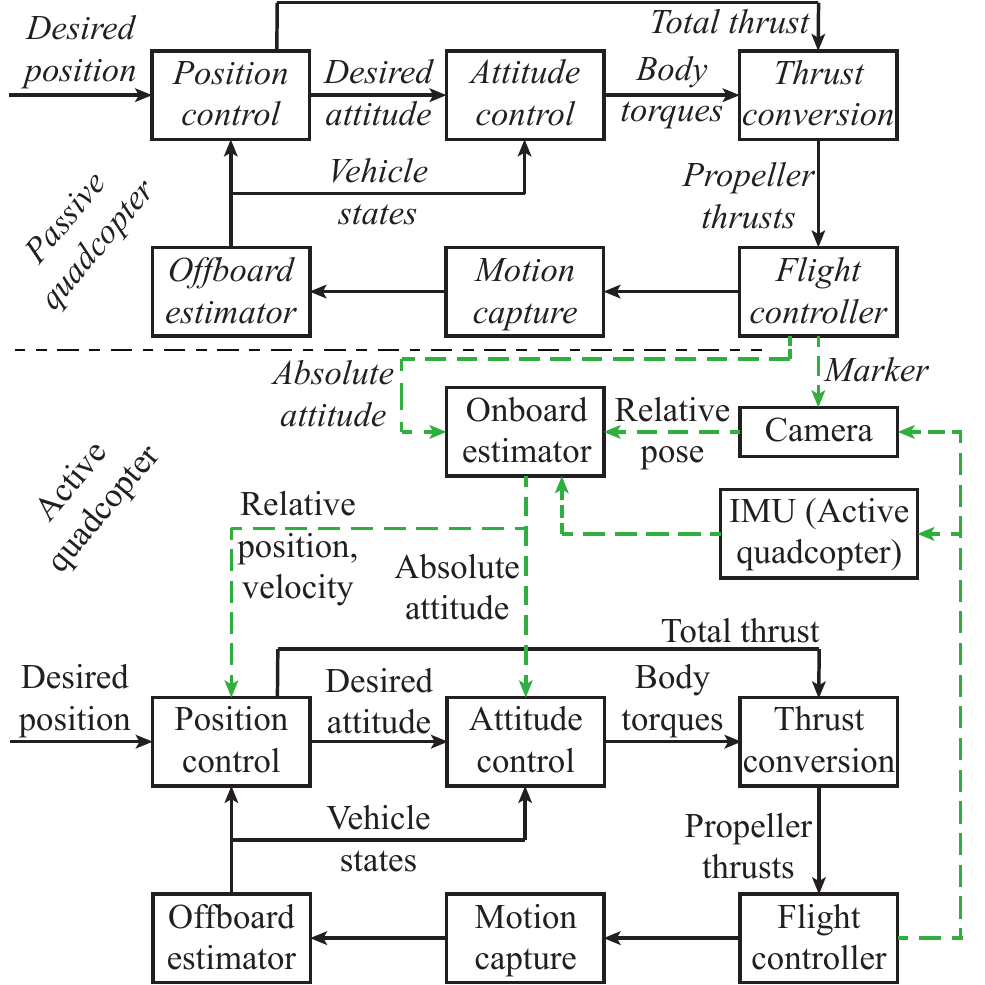}
	\vspace{-2mm}
	\caption{
		Control block diagram for the two quadcopters.
		All associations with the passive quadcopter are italicized.
		Black (solid) flow lines show the estimation and control scheme for both quadcopters using offboard sensing.
		Green (dashed) lines show the estimation scheme for the active quadcopter using onboard sensing, when the marker is in the FoV of the camera.
	}
	\vspace{-5mm}
	\label{fig:ControlDiagram}
\end{figure}

When the marker appears in the camera's FoV, we manually switch the active quadcopter to only use onboard sensing -- inertial measurements from an IMU and relative pose from vision.
The active quadcopter also rejects aerodynamic disturbances from the passive quadcopter's airflow by generating an additional thrust calculated from a feedforward map based on its relative position from the passive quadcopter.
The docking maneuver used is as presented in \cite{jain2020flying}.
The passive quadcopter continues using offboard sensing (from motion capture) for localization.
A block diagram for the described sensing and control scheme is shown in \figref{fig:ControlDiagram}.

\section{Relative State Estimation}\label{sec:relStateEst}

For estimation and control during midair docking, the states that we need to estimate are the relative position and velocity of the active quadcopter with respect to the passive quadcopter, and the absolute attitude of the active quadcopter.
So the states that we are interested in are $\position{\Q}{\F}{\E}, \velocity{\Q}{\F}{\E}, \R{\E}{\Q}$.
We follow the approach shown in \cite{mueller2018dynamics} to derive the prediction and measurement update steps.

\subsection{Prediction model}
We use the inertial measurements from the active quadcopter's IMU to predict the states forward in time.
Specifically, our estimator does not use the Newton-Euler equations of dynamics \eqref{eq:dynamics}.
This makes the estimator agnostic to the dynamics of the vehicles, and can work without the knowledge of unknown disturbance forces and torques.

Symbols with a tilde denote measurements.
The accelerometer measures the proper acceleration of the vehicle, and the measurement $\accelerometer{}$ is related to the vehicle's acceleration in the inertial frame as,
\begin{align}
	\accelerometer{\Q} &= \mrb{\R{\E}{\Q}}^{-1} \mrb{\acceleration{\Q}{\E}{\E} - \gravity{\E}} + \noiseAccel{\Q}
\end{align}
where $\noiseAccel{\Q}$ denotes the sensor noise.
We assume that the accelerometer is located at the vehicle's center of mass, so the measurement is unaffected by the vehicle's rotational motion.

The rate gyroscope measurement $\rateGyro{}$ is given as,
\begin{align}
	\rateGyro{\Q} &= \angVel{\Q} + \noiseGyro{\Q}
\end{align}
where $\noiseGyro{\Q}$ represents the sensor noise.

From the above equations, we derive the following,
\begin{equation}
\begin{aligned}
	\acceleration{\Q}{\E}{\E} &= \R{\E}{\Q} \accelerometer{\Q} + \gravity{\E} - \R{\E}{\Q} \noiseAccel{\Q} \\
	\angVel{\Q} &= \rateGyro{\Q} - \noiseGyro{\Q}
\end{aligned}
\end{equation}
which are used in the prediction step of the estimation.

The accelerometer and rate gyroscope are assumed to output regular measurements at high frequencies.

\subsection{Measurement model}
The camera ($\C$) is mounted rigidly on the active quadcopter ($\Q$) and the marker ($\M$) is pasted on the passive quadcopter ($\F$).
Therefore, the quantities $\R{\C}{\Q}$, $\R{\M}{\F}$, $\T{\Q}{\C}{\C}$, and $\T{\F}{\M}{\F}$ are fixed and known from the design.

Whenever the camera has the marker in its FoV, it outputs a relative position and orientation measurement of the marker with respect to the camera.
This can be written as a function of known quantities and system states as,
\begin{align}
    \cameraPos = - &\R{\C}{\Q} \transpose{\mrb{\rotation{\E}{\Q}}} \mrb{ \T{\Q}{\F}{\E} + \R{\E}{\F} \T{\F}{\M}{\F} } + \T{\Q}{\C}{\C} + \noiseCameraPos
	\label{eq:cameraPosMeas} \\
	\cameraAtt = &\R{\M}{\F} \transpose{\mrb{\R{\E}{\F}}} \R{\E}{\Q} \R{\Q}{\C} \expS{\noiseCameraAtt}
    \label{eq:cameraAttMeas}
\end{align}
where $\noiseCameraPos$ is the noise in relative position measurement, and $\noiseCameraAtt$ is the noise rotation vector \cite{pittelkau2003rotation} for relative attitude measurement.
The operation $\exp \mrb{\cdot} \colon \soThree \to \SOThree$ is the matrix exponential, and takes any element in the Lie algebra of the rotation group to the rotation group $\mrb{\SOThree}$ itself.

The relative pose measurements from the camera are available at a lower frequency than the inertial measurements.
The availability of these measurements is capped at the camera's frame rate, and also depends on marker detection which might be irregular.

\subsection{EKF implementation}
Following the method shown in \cite{mueller2017covariance}, we implement the extended Kalman filter (EKF) with a nine-dimensional state vector.
All estimated states' symbols are denoted with a hat.
The state vector consists of the active quadcopter's relative position $\relStatePos$ and relative velocity $\relStateVel$ with respect to the passive quadcopter, and a three-dimensional representation of the absolute attitude error $\absStateAttErr$, along with a reference orientation $\absStateAttRef$.
The resulting attitude is computed using,
\begin{equation}
	\absStateAtt = \absStateAttRef \expS{\absStateAttErr}
\end{equation}

For convenience of notation, we will represent the states as $\mrb{\relStatePos, \relStateVel, \absStateAttErr} = \mrb{\relStatePosEasy, \relStateVelEasy, \absStateAttErrEasy}$.
The estimated states are predicted forward in time using,
\begin{align}
	\relStatePosEasy_p \tPlus &= \relStatePosEasy \tNow + \relStateVelEasy \tNow \deltaT \\
	\relStateVelEasy_p \tPlus &= \relStateVelEasy \tNow + \nonumber\\
	&\mrb{\absStateAttRef\tNow \expS{\absStateAttErrEasy \tNow} \accelerometer{\Q} \tNow + \gravity{\E}} \deltaT \\
	\absStateAttErrEasy_p \tPlus &= \absStateAttErrEasy \tNow + \mrb{\rateGyro{\Q} \tNow - \smallHalf \skewMat{\rateGyro{\Q} \tNow} \absStateAttErrEasy \tNow} \deltaT \label{eq:absAttPred}
\end{align}
where the IMU data is assumed to be constant over the sampling time $\deltaT$.
Note that the attitude error is set to zero after each step, which makes \eqref{eq:absAttPred} valid.

The covariance of the estimated state $\covRelState \in \realMats{9}{9}$ is updated using the standard EKF equations,
\begin{equation}
	\covRelStatePred \tPlus = \mathbf{A}\tNow \covRelState\tNow \transpose{\mathbf{A}\tNow} + \mathbf{W}
\end{equation}
where the matrix $\mathbf{A}\tNow$ is given by,
\begin{equation}
	\mathbf{A}\tNow = \mmat{\identity	& \identity\deltaT	& \zeroMat \\
								\zeroMat	& \identity	& \skewMat{\absStateAttRef\tNow \accelerometer{\Q}\tNow} \deltaT \\
								\zeroMat	& \zeroMat	& \identity - \smallHalf \skewMat{\rateGyro{\Q}\tNow} \deltaT}
\end{equation}
where $\identity$ is the identity matrix and $\mathbf{W} = \mathrm{diag}(\zeroMat, \noiseVarAccel^2\identity, \noiseVarGyro^2\identity)$ consists the (isotropic) noise variances of the accelerometer and rate gyroscope respectively.

We set $t \leftarrow t+\Delta t$ for the remainder of the section for ease of notation.
We perform a post-prediction update to the state and covariance to reset the attitude error state to zero,
\begin{equation}
\begin{aligned}
	\relState_{p_+} &\tNow = \mrb{\relStatePosEasy_p \tNow, \relStateVelEasy_p \tNow, \zeroMat} \\
	\absStateAttRef_{p_+} &\tNow = \absStateAttRef \tNow \expS{\absStateAttErrEasy_p \tNow} \\
	\covRelStatePredPlus &\tNow = \TMat{\absStateAttErrEasy_p \tNow} \covRelStatePred \tNow \transpose{\mrb{\TMat{\absStateAttErrEasy_p \tNow}}} \label{eq:covAlignmentRelAtt}
\end{aligned}
\end{equation}
where, to keep the covariance aligned with the reference attitude, we use
\begin{equation}
	\TMat{\bs{\delta}} = \mathrm{diag} \mrb{\identity, \identity, \exp \mrb{-\smallHalf \skewMat{\bs{\delta}}}}.
\end{equation}

We then set $\relState\tNow = \relState_{p_+}\tNow$, $\absStateAttRef\tNow = \absStateAttRef_{p_+}\tNow$, and $\covRelState\tNow = \covRelStatePredPlus\tNow$.
If a relative pose measurement is available, we use the standard EKF formalism \cite{simon2006optimal} to perform the measurement update.
We assume that the absolute attitude of the passive quadcopter $\R{\E}{\F}$ is given to us.

We use \eqref{eq:cameraPosMeas} and \eqref{eq:cameraAttMeas} to derive the update equations.
We define a reference relative orientation,
\begin{equation}
	\cameraAttRef = \R{\M}{\F} \transpose{\mrb{\R{\E}{\F}}} \absStateAttRef \transpose{\mrb{\R{\C}{\Q}}}
	\label{eq:cameraAttRef}
\end{equation}
which is essentially, our best guess for the relative orientation measurement that we would get.

We now write the actual relative orientation measurement in terms of the reference measurement, and an attitude measurement error vector $\camAttError$ (which contains measurement noise) as,
\begin{equation}
	\cameraAtt = \cameraAttRef \expS{\camAttError}
	\label{eq:camAttError}
\end{equation}

We substitute \eqref{eq:cameraAttMeas} and \eqref{eq:cameraAttRef} in \eqref{eq:camAttError} to write,
\begin{align}
	\expS{\camAttError} &= \R{\C}{\Q} \expS{\bs{\delta}} \transpose{\mrb{\R{\C}{\Q}}} \expS{\noiseCameraAtt} \nonumber \\
	&= \expS{\R{\C}{\Q} \bs{\delta}} \expS{\noiseCameraAtt}
\end{align}

Note that the above is an exact expression, but cannot be simplified further.
We linearize it about $\bs{\delta}=\zeroMat$, and $\noiseCameraAtt=\zeroMat$,
\begin{equation}
	\camAttError = \R{\C}{\Q} \bs{\delta} + \noiseCameraAtt
\end{equation}

We consider $\camAttError$ as our measurements which can be derived from the actual relative orientation measurement and the reference value from \eqref{eq:camAttError} as,
\begin{equation}
	\camAttError\tNow = \rotInv{\transpose{\cameraAttRef}\!\!\!\tNow \; \cameraAtt\tNow}
\end{equation}
where $\rotInv{\cdot} \colon \SOThree \to \realVecs{3}$ is an operation that takes a rotation group element as input and outputs the corresponding rotation vector \cite{pittelkau2003rotation} for that element.
It is the inverse of the $\expS{\cdot}$ operation.

The EKF equations are derived by linearizing the measurements with respect to the state,
\begin{align}
	&\HMat = \frac{\partial \transpose{\mmat{\transpose{\cameraPos\tNow} & \transpose{\camAttError\tNow}}}}{\partial \mathbf{x}} \bigg|_{\mathbf{x}=\relState\tNow} \nonumber \\
	&= \mmat{ -\R{\C}{\Q} \transpose{\absStateAttRef}\!\!\!\tNow & \zeroMat & \R{\C}{\Q} \transpose{\absStateAttRef}\!\!\!\tNow \skewMat{\relStatePosEasy\tNow + \R{\E}{\F} \T{\F}{\M}{\F}} \\
	\zeroMat & \zeroMat & \R{\C}{\Q} }
\end{align}
The relative position measurement update is then done as,
\begin{equation}
\begin{aligned}
	\KMat &= \covRelState\tNow \transpose{\HMat} \big(\HMat \covRelState\tNow \transpose{\HMat} \\
	& \qquad \qquad \qquad \qquad +\mathrm{diag} \mrb{\noiseCovCameraPos, \noiseCovCameraAtt}\big)^{-1} \\
	\relState_{m}\tNow &= \relState\tNow + \KMat \innovation\tNow \\
	\covRelStateMeas\tNow &= \mrb{\identity - \KMat \HMat} \covRelState\tNow \\
	\absStateAttRef_{m}\tNow &= \absStateAttRef\tNow
\end{aligned}
\end{equation}
where $\noiseCovCameraPos$ and $\noiseCovCameraAtt$ are the covariance matrices of relative position measurement and relative orientation measurement, estimated experimentally.
The innovation $\innovation\tNow$ is given by,
\begin{equation}
	\innovation\tNow \! = \! \mmat{
		\cameraPos\tNow + \R{\C}{\Q} \transpose{\absStateAttRef}\!\!\!\tNow \mrb{\relStatePosEasy\tNow + \R{\E}{\F} \T{\F}{\M}{\F}} - \T{\Q}{\C}{\C} \\
		\camAttError\tNow}
\end{equation}

We employ a final correction step to reset the attitude error and update the reference orientation and state covariance,
\begin{equation}
\begin{aligned}
	\relState\tNow &= \mrb{\relStatePosEasy_{m}\tNow, \relStateVelEasy_{m}\tNow, \zeroMat} \\
	\absStateAttRef\tNow &= \absStateAttRef_{m}\tNow \expS{\absStateAttErrEasy_{m}\tNow} \\
	\covRelState\tNow &= \TMat{\absStateAttErrEasy_{m} \tNow} \covRelStateMeas \tNow \transpose{\mrb{\TMat{\absStateAttErrEasy_{m} \tNow}}}
\end{aligned}
\end{equation}

The state estimator then continues to absorb the next set of IMU measurements.

\section{Hardware Design}\label{sec:hardwareDesign}
In this section, we explain the design of the midair docking mechanism, the quadcopters, and the vision system used in our experiments.

\subsection{Docking mechanism}
We use a midair docking mechanism whose design and working has been demonstrated in \cite{jain2020flying}.
This design uses a mechanical guide structure in the form of a docking platform on the active quadcopter, and docking legs on the passive quadcopter as can be seen in \figref{fig:FilmReelDocking}.

Particularly, this mechanism does not use active components which allows for low weight and complexity.
It also allows for a margin of error of a few centimeters during the docking process.
Lastly, the quadcopters dock vertically aligned with this mechanism.
This enables an easy undocking process -- the passive quadcopter can do a regular take-off from the docking platform.

\subsection{Vehicle design}
We use Crazyflie 2.0 flight control boards \cite{giernacki2017crazyflie} for both quadcopters.
\figref{fig:FilmReelDocking} shows pictures of the two vehicles.

\subsubsection{Active quadcopter}
The active quadcopter is designed to have sufficient payload capacity to carry useful sensors such as surveillance cameras or environmental sensors.
It uses a 3S \SI{2.2}{\ampereHour} lithium-ion polymer (LiPo) battery to power itself, and weighs \SI{825}{\gram} including the battery.
Its arm length is \SI{165}{\milli \meter} and it uses propellers of diameter \SI{203}{\milli \meter}.
A lightweight camera is mounted on the quadcopter, facing upwards, to allow onboard sensing and localization.
The docking platform is stacked on top of the quadcopter, and has a hole in it for the camera to have a clear view.

\subsubsection{Passive quadcopter}
The passive quadcopter is designed to have enough payload capacity to carry a package of about \SI{150}{\gram}.
The docking legs are attached below each of the motors to minimize propeller airflow blockage, so that payload capacity is not affected.
It uses a 2S \SI{0.8}{\ampereHour} LiPo battery to power itself, and weighs \SI{160}{\gram} including the battery.
Its arm length is \SI{58}{\milli \meter} and it uses propellers of diameter \SI{76}{\milli \meter}.
It carries a marker that is pasted below its frame which can be detected by a camera.

\subsection{Vision system}
To perform a relative vision-based localization, we require a camera, a marker which can be easily detected by the camera, and an algorithm that can extract position and orientation information from the marker image.
For this purpose, we use ArUco marker \cite{garrido2016generation, romero2018speeded},
part of the OpenCV \cite{opencv_library} Contrib library.
To see and detect the markers, we use a JeVois-A33 Smart Machine Vision Camera \cite{nair2020performance}.
It is light-weight (\SI{17}{\gram}), has a \SI{60}{\degree} FoV, and is equipped with an onboard processor which can run the ArUco detection algorithm at \SI{30}{\hertz}.

Since the multirotors fly in proximity during the docking process, it is possible for a part of the marker to go outside the FoV of the camera, due to movements caused by aerodynamic disturbances.
When this happens, the algorithm does not detect the marker.
To mitigate this, we embed a smaller ArUco marker inside the original marker pasted on the passive quadcopter, which is verified to always remain in the FoV of the camera via repeated docking experiments.

The lighting condition for a camera facing upwards looking at a marker facing downwards are not favorable in indoor or outdoor flight, since light sources are usually on ceilings in indoor environments, and in the sky in outdoor settings.
To improve marker detection, retroreflective tape is pasted on the white portions of the marker.
An LED is attached near the camera's lens.
This enables more robust marker detection because of the retroreflected LED light from the marker.

\section{Experimental Validation}\label{sec:expValidation}
The feasibility of using relative vision to localize a vehicle with respect to another for a sensitive docking maneuver is verified experimentally.
This section covers characterization of sensor noise standard deviations, the overall system setup for the experiments, and finally some results from the docking experiments using relative vision.

\subsection{Sensor Noise Characterization for EKF}
The flight controller on the vehicles has an IMU which returns accelerometer and rate gyroscope data at \SI{500}{\hertz}.
We estimate the sensor noise standard deviations for the IMU, $\noiseVarAccel$ and $\noiseVarGyro$, from in-flight data.
\begin{equation}
	\noiseVarAccel = \SI{0.5}{\meter \per \second \squared} \quad \text{and} \quad \noiseVarGyro = \SI{0.1}{\radian \per \second}
\end{equation}

The noise covariance in the relative position and orientation measurements are estimated from static tests where the camera and marker are placed at various fixed positions with respect to each other.
The marker side length is \SI{42}{\milli \meter}.
The relative orientation noise covariance remained approximately constant for the various separations.
The relative position noise covariance was found to be dependent on the distance along the optical axis $z$ between the camera and the marker.
Specifically, the values were estimated to be,
\begin{align}
	\noiseCovCameraPos &= \mathrm{diag} \mrb{0.2^2, 0.2^2, 0.3^2} \mrb{\frac{z}{\SI{1.0}{\meter}}}^2 \SI{}{\m \squared} \\
	\noiseCovCameraAtt &= \mathrm{diag} \mrb{0.35^2, 0.35^2, 0.05^2} \SI{}{\radian \squared}
\end{align}
We would like to note that the off-diagonal terms were not exactly zero, but were much smaller than the diagonal terms and hence neglected.

We see that the closer the marker is with respect to the camera, the lower is the position measurement noise covariance.
We also note that the position measurement performance of the camera is better perpendicular to the optical axis, than along it.
This is because the horizontal position is calculated from the position of the marker in the image plane, whereas the vertical position (along the optical axis) is calculated from the size of the marker in the image plane.
Similarly, the orientation measurements are more precise about the optical axis as it is directly seen as a rotation in the image plane, whereas the other two orientation measurements are calculated from the projected side lengths of the marker in the image plane.
The camera measures the relative pose of the marker at approximately the frame rate, which is set to \SI{30}{\hertz}.
The marker does go undetected in certain frames due to motion blur from vibrations and fast relative motion.
However, this irregularity is not found to affect the localization and docking performance much.

\subsection{Experimental Setup}
The motion-capture system measures the absolute position and orientation of each of the vehicles and runs at \SI{200}{\hertz}.
The flight controller of the active quadcopter is connected to the camera via a serial port (UART) and transmits IMU data to the camera at \SI{500}{\hertz}.
The EKF is run on the camera's processor at \SI{500}{\hertz}, predicting the states forward using the IMU data, and doing the relative pose measurement updates at around \SI{30}{\hertz} based on the marker detection.
The passive quadcopter's attitude is sent to the active quadcopter via a radio at approximately \SI{100}{\hertz}, which is used in the EKF.
The state estimates are sent from the camera to the flight controller at \SI{500}{\hertz}, which are then used for closed-loop control.
During closed-loop control of the active quadcopter via onboard sensing, we continue recording motion capture measurements as ground truth, to evaluate the performance of the onboard state estimator.

\subsection{Results}
The experiment that we conduct involves the active quadcopter flying to a particular location, using offboard sensing, and the passive quadcopter flying to about \SI{0.6}{\meter} vertically above the active quadcopter, again using offboard sensing.
Once the passive quadcopter's marker is in the FoV of the active quadcopter's camera, we close the loop and start using purely onboard sensing for flying the active quadcopter.
A docking command is then issued to the active quadcopter where it starts to ascend and close the vertical gap between itself and the passive quadcopter.
Once the docking platform is within \SI{15.0}{\centi \meter} vertically from the docking legs of the passive quadcopter, and their geometric centers are within \SI{2.5}{\centi \meter} in the horizontal plane (based on the onboard estimate), the passive quadcopter is commanded to stop its motors and fall freely onto the docking platform.
The docking platform is forgiving to small position and yaw errors and thus the passive quadcopter slides into the platform and docks with the active quadcopter.
Simultaneously, the active quadcopter automatically switches back to using offboard sensing, since the marker is no longer visible at such a close distance, and updates its mass and moment of inertia.

For undocking, the passive quadcopter is simply commanded to start hovering at the previous desired position, and the active quadcopter is commanded to go vertically downwards by the original vertical separation.

\begin{figure}
	\centering
	\includegraphics[width=\columnwidth]{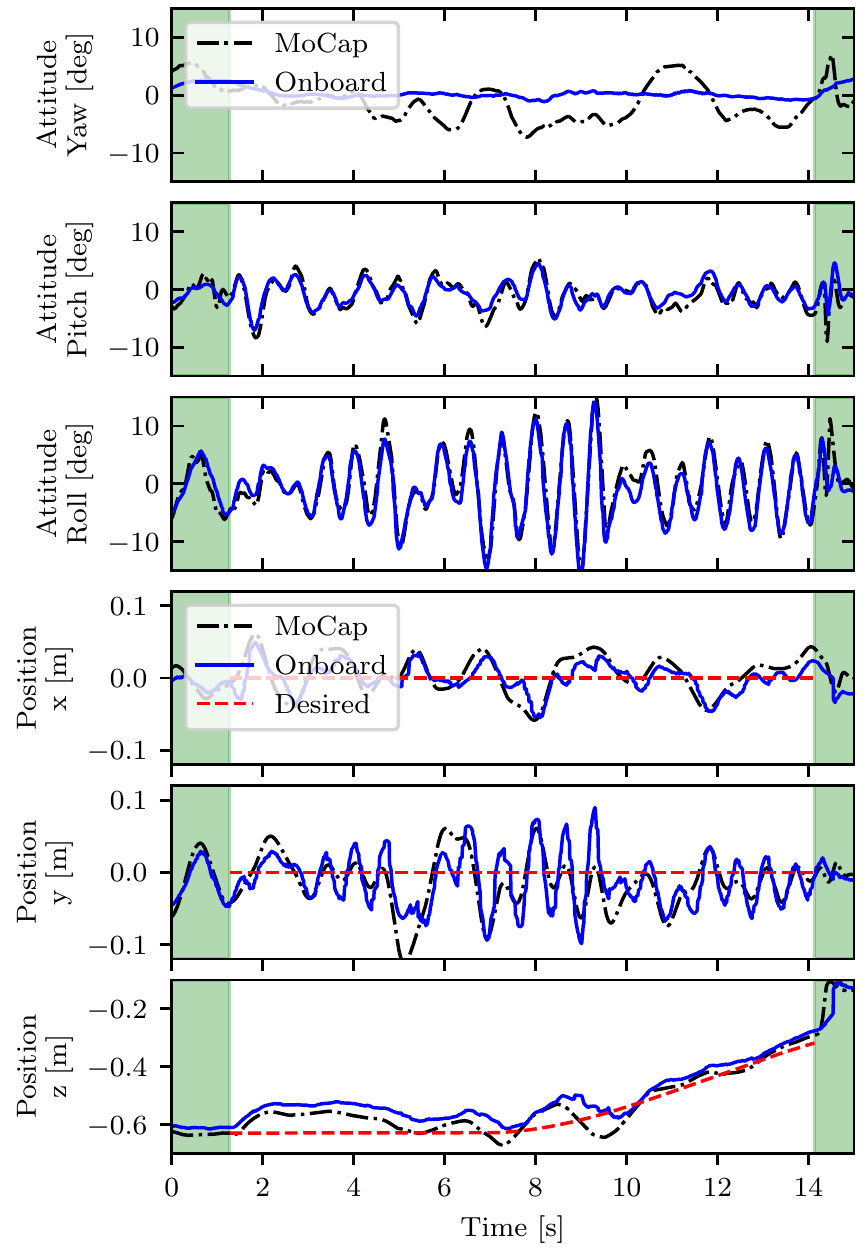}
	\vspace{-7mm}
	\caption{
		Attitude and relative position estimates from the onboard state estimator compared with the motion-capture values.
		We also plot desired position.
		Green background (at the start and end of the plot) means that the active quadcopter is controlled using offboard state estimates, while clear background means that onboard state estimates are used.
		The position estimation error is found to be lesser than \SI{2}{\centi \meter} when the two quadcopters are in docking range.
		At about \SI{14}{\second}, the vehicles dock.
	}
	\vspace{-5mm}
	\label{fig:estimateComparison}
\end{figure}

The experiment was verified for repeatability by successfully performing docking using onboard sensing five times.
The experimental footage can be seen in the video attachment.
\figref{fig:FilmReelDocking} shows a sequence of snapshots of the docking procedure.
We can see that the marker is detected even when the two quadcopters are flying very close (a green border is shown around the detected marker), but remains undetected in the docked configuration.

We show plots of estimated position and attitude from the EKF, and the motion-capture values for a sample experiment during the docking maneuver in \figref{fig:estimateComparison}.

The estimation error (difference between onboard estimate and ground truth) in position is always within \SI{10}{\centi \meter}, and reduces to less than \SI{2}{\centi \meter} when the vehicles get close enough for docking.
The attitude estimation error for yaw is within \SI{5}{\deg} which is noticeably higher than the estimation error in roll and pitch.
The reason for this is that the yaw rotation (about gravity) is mostly observed purely via the marker, whereas the roll and pitch rotations are observed via both the marker and the accelerometer measurement which remains mostly nearly parallel to gravity during flight.

\section{Conclusion}\label{sec:conclusion}
We have developed a relative state estimation algorithm based on EKF formulation to localize one ``active'' multirotor with respect to another ``passive'' multirotor, using just onboard sensors.
Specifically, the active multirotor uses its accelerometer, rate gyroscope, and an onboard camera looking at a marker on the passive multirotor to determine its relative position and velocity with respect to the passive multirotor, and its absolute orientation with respect to an inertial frame.
Notably, the estimator requires no knowledge of the physical parameters of the vehicles such as mass, moment of inertia, or aerodynamic properties.
We also presented a control architecture that can use these relative state estimates to control the active quadcopter at a desired setpoint with respect to the passive quadcopter.
The performance of estimator and control architecture was experimentally validated via repeatable midair docking experiments where we were able to successfully dock the multirotors -- a task requiring position precision on the order of a centimeter.

A possible extension to this work is to use onboard sensing for absolute localization of both the passive and active quadcopter, for example via GPS \cite{nemra2010robust}.
This will allow extension of this work's contribution to outdoor applications of modular and cooperative robots.

\section*{Acknowledgment}
We gratefully acknowledge financial support from NAVER LABS and Code42 Air.

The experimental testbed at the HiPeRLab is the result of contributions of many people, a full list of which can be found at \url{hiperlab.berkeley.edu/members/}.

\balance
\bibliographystyle{IEEEtran}
\bibliography{main}

\end{document}